\documentclass[letterpaper, 10 pt, conference]{ieeeconf}  

\IEEEoverridecommandlockouts                              

\usepackage{comment}
\usepackage{times}
\usepackage{epsfig}
\usepackage{graphicx}
\usepackage{amsmath}
\usepackage{amssymb}
\usepackage{cite}
\usepackage{xcolor}
\usepackage{multirow}
\usepackage{booktabs}
\usepackage{array}
\usepackage{algorithm}
\usepackage[noend]{algpseudocode}
\usepackage{makecell}
\usepackage{paralist}
\graphicspath{{imgs/}}
\usepackage{soul}
\usepackage{caption}
\usepackage{pifont}
\usepackage{hyperref}
\usepackage{subcaption}
\usepackage[font=footnotesize]{caption} 
\makeatletter
\g@addto@macro\normalsize{%
  \setlength\abovedisplayskip{4pt}
  \setlength\belowdisplayskip{4pt}
  \setlength\abovedisplayshortskip{4pt}
  \setlength\belowdisplayshortskip{4pt}
}
\makeatother

\DeclareMathOperator*{\argmax}{arg\,max}

\newcommand{\mymat}[1]{\mathbf{#1}}

\newcommand{\ours}{VioLA}

\newcommand{\figref}[1]{Fig.~\ref{#1}}

\title{\LARGE \bf
VioLA: Aligning Videos to 2D LiDAR Scans
}

\author{Jun-Jee Chao$^{\dagger}$, Selim Engin$^{\dagger}$, Nikhil Chavan-Dafle, Bhoram Lee, and Volkan Isler
\thanks{\small{$\dagger$ indicates equal contribution.}}
\thanks{\small{All authors are with Samsung AI Center NY, USA. Email:}}
\thanks{\small{junjee.chao@partner.samsung.com, kazim.engin@samsung.com}}}%

\begin{document}

\maketitle
\thispagestyle{empty}
\pagestyle{empty}

\begin{abstract}
We study the problem of aligning a video that captures a local portion of an environment to the 2D LiDAR scan of the entire environment. We introduce a method (\ours{}) that starts with building a semantic map of the local scene from the image sequence, then extracts points at a fixed height for registering to the LiDAR map. Due to reconstruction errors or partial coverage of the camera scan, the reconstructed semantic map may not contain sufficient information for registration. To address this problem, \ours{} makes use of a pre-trained text-to-image inpainting model paired with a depth completion model for filling in the missing scene content in a geometrically consistent fashion to support pose registration. We evaluate \ours{} on two real-world RGB-D benchmarks, as well as a self-captured dataset of a large office scene. Notably, our proposed scene completion module improves the pose registration performance by up to $20\%$.
\end{abstract}

\section{Introduction}
\label{sec:intro}
Generating 3D semantic maps of home environments enables numerous applications in immersive technologies, home-robotics, and real estate. Even though commercial grade solutions exist for generating 3D maps of home environments, they require expensive and specialized hardware, as well as meticulous scanning procedures. Alternatively, using widely available cameras such as those on mobile phones can be used to reconstruct a local area. However, scanning an entire house with a single camera is difficult; getting all details in one scan is tedious even for a single room. Moreover, merging scans across rooms is error-prone due to the lack of images with overlapping features.
In this paper, we present a method to overcome these challenges using a 2D floor layout, such as those obtained by Robot Vacuum Cleaners (RVCs), and user-scanned videos recorded \emph{independently} with an RGB-D camera.



\begin{figure}[t]
    \centering
    \includegraphics[width=\columnwidth]{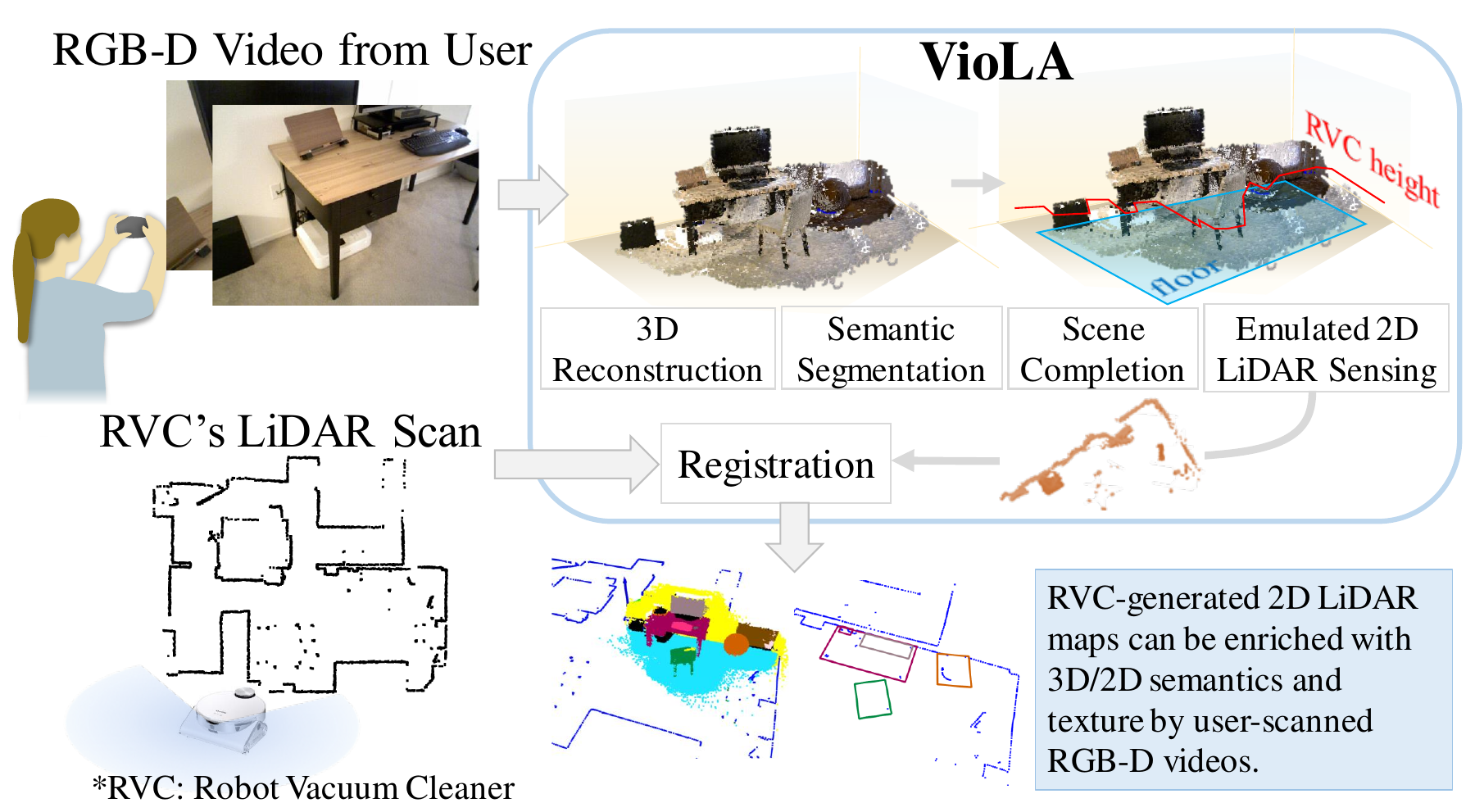}
    \caption{Given an RGB-D image sequence from an indoor scene and a 2D LiDAR scan of the same environment, our task is to align the video to the LiDAR map by registering the first camera pose to the LiDAR coordinate frame. After registration, our method allows augmenting the LiDAR map with 3D geometry, texture and semantics information.}
    \label{fig:teaser}
    \vspace{-15pt}
\end{figure}

\begin{figure}[!t]
    \centering
    \includegraphics[width=.95\linewidth]{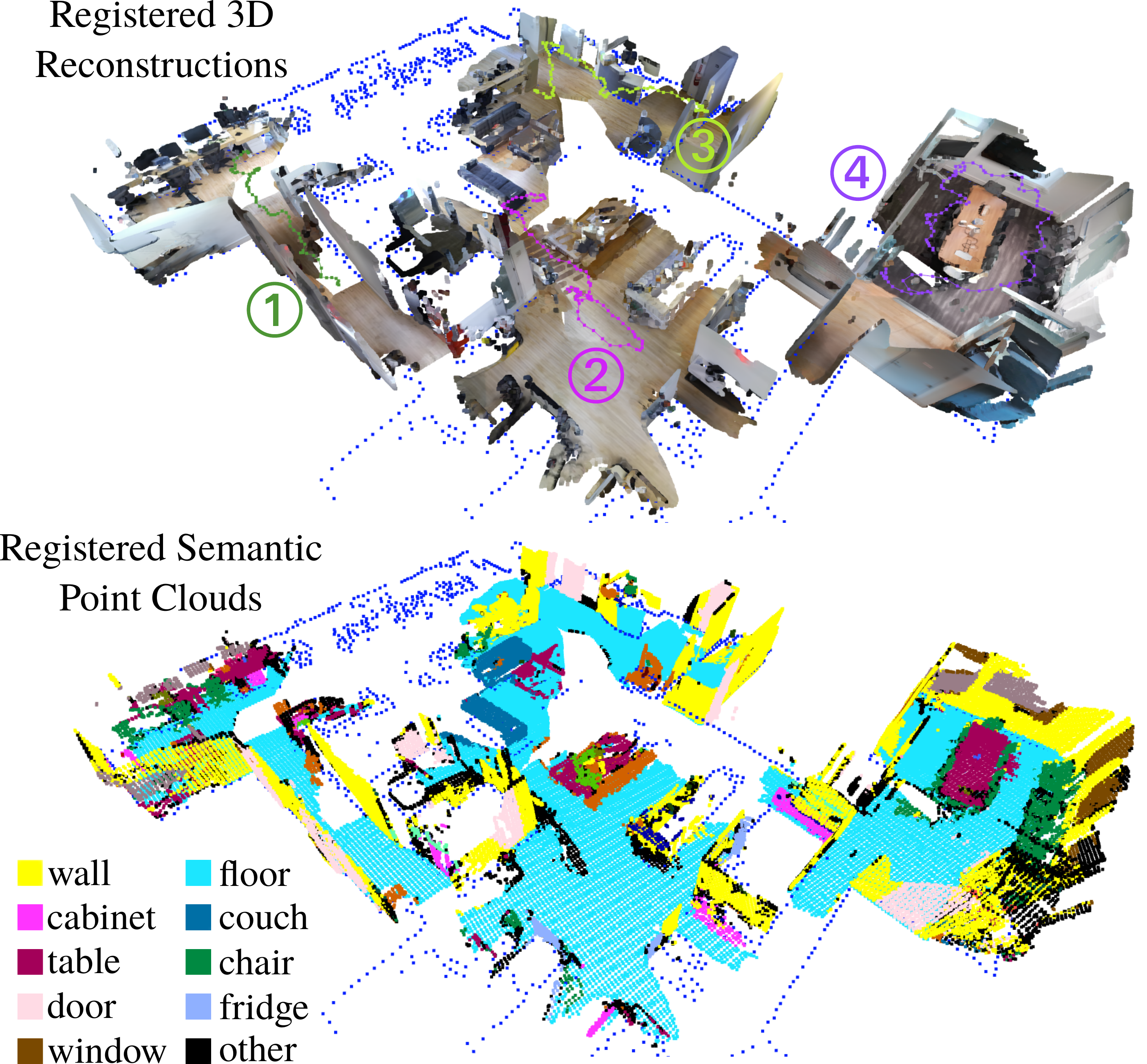}
    \caption{Using \ours{}, 3D reconstructions obtained over multiple scans (4, in this case) can be fused via registration to the LiDAR map (blue dots). In addition, our method can generate the semantic labels for the fused point cloud. The colored lines in the top figure show the camera trajectories and the point colors in the bottom figure indicate the classes given in the legend.}
    \label{fig:office-scan}
    \vspace{-15pt}
\end{figure}

We study the following problem: Given a 2D LiDAR map of an environment and an RGB-D image sequence recorded from a local section of the same environment, the task is to align the pose of the first image to the LiDAR map as shown in~\figref{fig:teaser}.
The benefits of aligning videos to 2D LiDAR scans are twofold. First, image sequences with dense scene information can be used to augment 2D LiDAR maps with 3D geometry, texture and semantics information. This is useful for example to disambiguate walls and furniture in 2D LiDAR maps, which may allow for better experiences of user-robot interaction. 
Second, the LiDAR map can serve as a common coordinate frame to align short clips of videos captured from different locations of the same house. Therefore, the entire house can be scanned by aligning independent video sequences to the common LiDAR map, for example, as seen in \figref{fig:office-scan}.

Currently, there is no existing solution for the problem of registering raw 2D point clouds obtained by LiDAR measurements to 3D reconstructions from RGB-D image sequences. This alignment task poses unique challenges in several aspects. In particular, point clouds from 2D LiDAR scans contain information from the entire floor plan of an apartment but only at a fixed height, with no semantic context. Whereas reconstruction point clouds have denser 3D information but only from a local area. Therefore, neither point set is a superset of the other, which makes matching or registration challenging. Furthermore, the sensing modalities are different and the representation to use for registering a 3D reconstruction to a LiDAR map is not trivial.




We propose \ours{} (Video-LiDAR Alignment), a method for aligning videos to 2D LiDAR maps. The underlying idea of \ours{} is to reconstruct the 3D from RGB-D and perform 2D ray casting to emulate LiDAR measurements of RVC. If the video scan captures majority of the scene at the RVC height, then we can use the ray cast hit points directly to register with the LiDAR point cloud. However, this may not be always the case. User-captured videos may not contain enough information of the scene at the RVC height leading to poor registration performances. To fill in these missing regions of the reconstruction, \ours{} leverages a pre-trained text-to-image model to synthesize images at novel viewpoints depending on the scene geometry and lifts them to 3D with a depth completion module in a geometrically consistent fashion. Then, the completed point cloud can be used to perform 2D ray casting at the RVC height and 2D-2D point cloud registration with respect to the LiDAR map.
The contributions in this paper can be summarized as follows.

\begin{itemize}
    \item We introduce the novel task of registering videos to 2D LiDAR scans, which is motivated by \emph{i}) enriching LiDAR maps with texture and semantics for better user experience, and \emph{ii}) making use of LiDAR maps as a common frame to localize short clips of videos taken from different locations in an indoor environment.
    \item We present \ours{}: A technique to register 3D reconstructions of local areas to 2D LiDAR maps captured from indoor environments such as apartments.
    \item As part of \ours{}, we design a strategy for viewpoint selection and leverage a state-of-the-art text-to-image model to synthesize images at locations with missing information to aid our point cloud registration module.
    \item We evaluate \ours{} on two benchmarks using real videos and synthetic LiDAR maps, in addition to a self-collected dataset of real videos and LiDAR scans. 
\end{itemize}

\section{Related Work}
In this section, we provide an overview of existing work that is related to our problem setup.

\textbf{Cross-modality registration.}
There is a line of research that studies the problem of cross-modality registration between 2D images and 3D point clouds.~\cite{yu2020monocular} uses 2D-3D line correspondences to estimate the camera pose. Other recent works apply deep learning to register 2D images to 3D point clouds~\cite{Feng2019ICRA,li2021deepi2p,ren2022corri2p,miao2023poses,jeon2022efghnet,CHEN2022209}. However, none of these methods directly addresses the problem of registering videos to 2D point clouds.
More similar to our setting,~\cite{wijmans2017exploiting} and~\cite{sokolova2022iros} make use of floor plan images for aligning RGB-D scans. While it operates on building-scale floor plans, the work in~\cite{wijmans2017exploiting} requires RGB-D panorama images that cover a large portion of the floor plan. On the other hand,~\cite{sokolova2022iros} takes as input an RGB-D sequence that scans the entire indoor scene to refine the camera poses, which can be a tedious task for end users.

\textbf{Point cloud registration.}
Our problem can be formulated as point cloud registration by first reconstructing the 3D scene from the video. However, it is still non-trivial to perform pose registration between 3D point clouds of local area and 2D LiDAR point clouds of the entire floor plan.

ICP~\cite{besl1992method,segal2009generalized} solves the registration problem by building correspondences between closest points, which is sensitive to initialization and prone to local minima. To overcome the initialization issue, FGR~\cite{zhou2016fast} and TEASER~\cite{Yang20tro-teaser, yang2019polynomial} optimize one-to-one correspondence-based object functions with the help of 3D point feature descriptors like PFH~\cite{rusu2008aligning} and FPFH~\cite{rusu2009fast}.
Instead of building one-to-one correspondences in a single shot,~\cite{chao2023category} proposes an optimization loss function that consider multiple correspondences at the initial stage.

Deep learning is also applied to learn point feature descriptors~\cite{deng2018ppfnet,zeng20173dmatch,gojcic2019perfect,wang2019deep,ginzburg2022deep}. PointNet~\cite{qi2017pointnet++} is a popular architecture to extract features directly from raw point clouds~\cite{yaoki2019pointnetlk, deng2018ppfnet}. Some other works apply DGCNN for feature learning~\cite{wang2019deep,wang2019prnet}. Recently, more works have been developed for partial-to-partial point cloud registration~\cite{yew2020-RPMNet,Fu2021RGM,jiang2021sampling,wu2021feature,idam,lee2021deeppro, predator, choy2020deep, qin2022geometric}. However, most of them focus on 3D-to-3D registration, which are not directly applicable to the 3D-to-2D case studied in this paper.

\textbf{Point cloud completion.}
There has been a lot of attention on the task of object-level point cloud completion~\cite{zhang2021view, yu2021pointr, li2021spgan, yan2022shapeformer, hani20233d, autosdf2022}, for which priors for object shapes are learned from data. Scene-level completion methods have been proposed with a focus on self-driving car datasets~\cite{li2023voxformer, xia2023scpnet}. Other works that focus on single-view scene completion~\cite{song2017semantic,cai2021semantic, cao2022monoscene,ha2022semabs,wang2023semantic} are usually limited to scene completion only in the field of view. However, completing the scene in the field of view is not enough for our task, as we need to fill in the missing regions that are not seen by the cameras to aid pose registration. 
There are recent methods that utilize text-to-image inpainting models and depth estimation models to generate scenes given text prompts~\cite{hoellein2023text2room, SceneScape}. 
However, they require predefined camera trajectories, and the synthesized 3D content is not grounded to any actual scene, which makes them difficult to use for pose registration purposes.  

\section{Method} 

\begin{figure*}
    \centering
    \includegraphics[width=.9\linewidth]{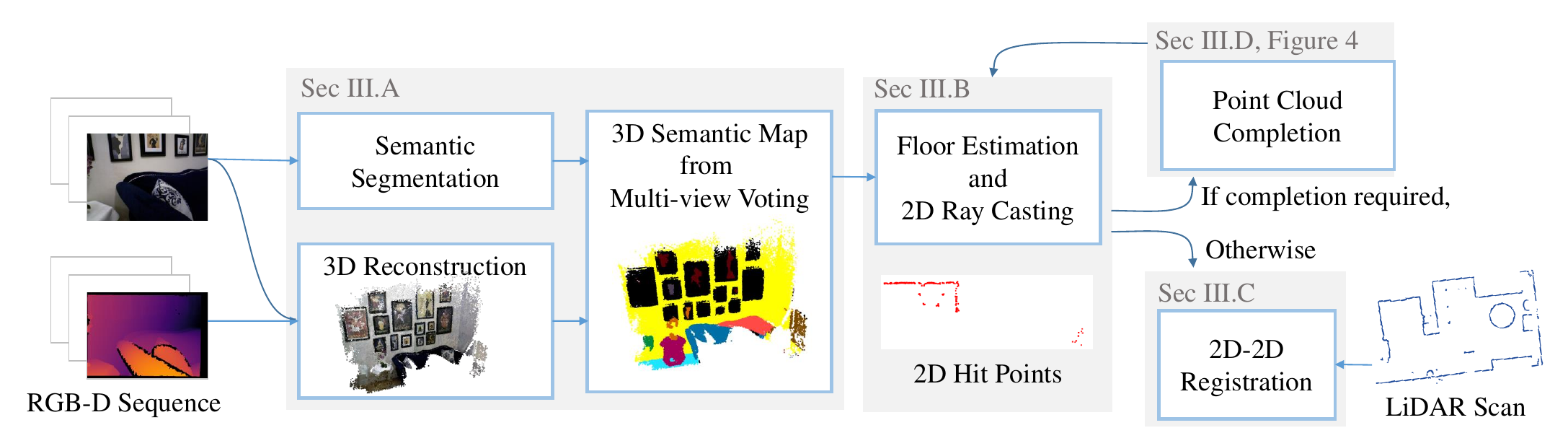}
    \caption{\textbf{Method Overview.} \ours{} takes as input an RGB-D sequence and a 2D LiDAR scan from uncalibrated sensors, and aligns these two sources of measurements. Our approach first reconstructs the 3D scene and extracts semantics from each image which are then fused by multi-view voting to obtain a semantic point cloud. After finding the floor surface, it performs 2D ray casting at a desired height to emulate LiDAR measurements. If the hit points of these ray casts cover a large portion of the reconstruction they are used directly to align to the input 2D LiDAR scan with our 2D-2D point cloud registration module. If not, \ours{} uses our novel strategy for view selection coupled with inpainting-based scene completion. The final result of our method is the $SE(2)$ transformation relating the first camera frame to the LiDAR coordinate frame, as well as the semantic map.}
    \label{fig:overview}
    \vspace{-15pt}
\end{figure*}

The input to \ours{} consists of an RGB-D video sequence $V = \{I_1, \dots, I_n\}$ and a 2D point cloud $\mymat{P} \in \mathbb{R}^{2 \times M}$ captured by the LiDAR sensor on a ground robot such as a robot vacuum cleaner (RVC).
The output is an estimate of the transformation $^{L} \mymat{T}_V \in SE(2)$ that aligns the pose of the first image in the video to the LiDAR map, as well as a 3D semantic map of the scene registered to the LiDAR coordinate frame. For brevity, we denote the pose estimates by $\mymat{T}$ in the remainder of the paper. The main module of \ours{} includes: a) 3D reconstruction and semantic segmentation from the RGB-D sequence, b) 2D point cloud extraction from (a) by floor estimation and ray casting, and c) 2D registration of point clouds from (b) and LiDAR point cloud from the RVC. Additionally, to address the case where the registration suffers from having too few points from ray casting, \ours{} utilizes d) a 3D scene completion module using a pre-trained inpainting model to synthesize the 3D geometry in missing regions.
An overview of our approach is shown in~\figref{fig:overview}. In the following subsections, we detail each of the modules.

\subsection{3D reconstruction, semantic segmentation}\label{sec:3D_semantic}

We use off-the-shelf SLAM algorithms~\cite{teed2021droid, choi2015robust, park2017colored} to reconstruct 3D point clouds from the RGB-D videos. 
We further estimate the segmentation class for each reconstructed 3D point by fusing 2D semantic segmentation into 3D. The semantic segmentation provides rich information about the scene including the floor which is critical for the next step (Sec. \ref{sec:ray_cast}).
Following~\cite{siddiqui2023panoptic}, we first apply 2D color-based augmentations on the key frames of the SLAM algorithm. These augmented images are then used to predict the semantic masks using Mask2Former~\cite{cheng2021mask2former}. These estimated masks of a single frame can then be fused into a categorical probability distribution $\textrm{p}_u=[p_u^{(1)},...,p_u^{(C)}]$ over $C$ classes for each pixel $u$ paired with a confidence score $\textrm{s}_u \in [0, 1]$. To determine the semantics of the reconstructed 3D points, we first project each point $\mymat{x}$ onto the $m$ SLAM camera frames in which the point is visible and store the set of pixels $\mathcal{U} = \{u_j\}_{j=1}^m$ that $\mymat{x}$ projects to. Then, we sum over the corresponding probability distribution weighted by the normalized confidence scores. The final semantic label $c(\mymat{x})$ of the point $\mymat{x}$ is determined as: 
\begin{equation}
c(\mymat{x}) = \argmax_{i \in \{1, \dots, C\}} \sum_{u \in \mathcal{U}}  w_u \cdot p_u^{(i)}
\label{eq:class}
\end{equation}
where $w_u = \textrm{s}_u / \sum_{v  \in \mathcal{U}} \textrm{s}_v$ is the normalized confidence score, and $u = \Pi (\mymat{x})$ is the projection of $\mymat{x}$ in each of the $m$ frames where $\mymat{x}$ is visible.


\subsection{Floor normal estimation and RVC viewpoint projection}\label{sec:ray_cast}

After reconstructing the scene from the RGB-D video, we simulate what the RVC would see in this scene in order to perform 2D point cloud registration with the LiDAR map. To do so, we first estimate the floor surface by fitting a plane to the points that are labeled as floor. The ground is assumed to be visible in the video, which is usually the case in casually recorded videos as shown in the experiments section. Then we move all SLAM camera poses down to a predefined RVC height while preserving only the yaw angles, i.e., the $y$-axis of the camera is parallel to the estimated floor normal. From these downprojected camera poses, we cast rays from the camera centers to the reconstructed 3D points to emulate LiDAR hit points.    

\subsection{Pose initialization and optimization}\label{sec:pose_optim}

Since SLAM reconstructs the 3D points with respect to the first camera frame, the hit points $\mymat{H} \in \mathbb{R}^{2 \times N}$ obtained from the 2D ray casting module are also in the same frame. \ours{} performs 2D-2D point cloud registration between the LiDAR map $\mymat{P}$ and the ray cast hit points $\mymat{H}$.
Similar to most iterative algorithms for registration, our method requires an initial guess for estimating the relative pose between the two point clouds. We rasterize both the LiDAR map and the simulated hit points into images. We then perform template matching and select the top $k$ poses with the highest normalized cross correlation scores as our initial poses. With these initial poses, we apply GPU parallelization to simultaneously update the poses by minimizing the following loss function using gradient descent:
\begin{equation} 
\mathcal{L} = \sum_{i=1}^k d( \mymat{T}_i  \cdot \mymat{H}, \mymat{P})
\end{equation}
where $d(\cdot, \cdot)$ denotes the one-directional Chamfer distance given by $d(X, Y) = \frac{1}{|X|} \sum_{\mathbf{x} \in X} \min_{\mathbf{y} \in Y} ||\mathbf{x} - \mathbf{y}||_2$, and the point clouds are in homogeneous coordinates. We compute the Chamfer distance from the hit points to the LiDAR map after applying the estimated transformations. Finally, the estimated pose is selected as the pose that minimizes the one-directional Chamfer distance after optimization. 

This Chamfer-based optimization can be replaced with other pose optimization methods like ICP. In a preliminary experiment, we found out that ICP performs very similarly to our method given the same initialization. However, ICP on average takes a total of $49.5$ seconds for matching with $k=100$ initializations, while our method takes $15.5$ seconds on average. Therefore, we use the Chamfer-based pose optimization.

\subsection{Scene completion based on image inpainting} \label{sec:view_select_complete}

As we will demonstrate later in experiments in Section~\ref{sec:exp_pose}, reconstructed points at the RVC height is critical for registration success. 
However, these points might be missing due to the video not capturing the lower part of the scene or the SLAM algorithm suffering from matching featureless points.
In this section, we show how state-of-the-art inpainting methods can be used to provide this missing information. Note that we are not trying to fill in every detail of the scene. Instead, we show how one can judiciously select good viewpoints so that \textit{i})~resulting images contain a sufficient amount of observed pixels so that inpainting is successful, and \textit{ii})~the inpainted pixels correspond to locations that would have been observed by the RVC. 

\textbf{Virtual viewpoint selection.}
The key idea of our strategy for viewpoint selection is to incrementally add 3D content from a sequence of virtual views. 
Our strategy involves starting from a view that sees the boundary between observed and unobserved pixels, then moving the camera back to increase the field of view, and finally rotating to cover more unobserved pixels.

\begin{figure}
    \includegraphics[trim={0 7.5cm 9.5cm 0cm},clip, width=\linewidth]{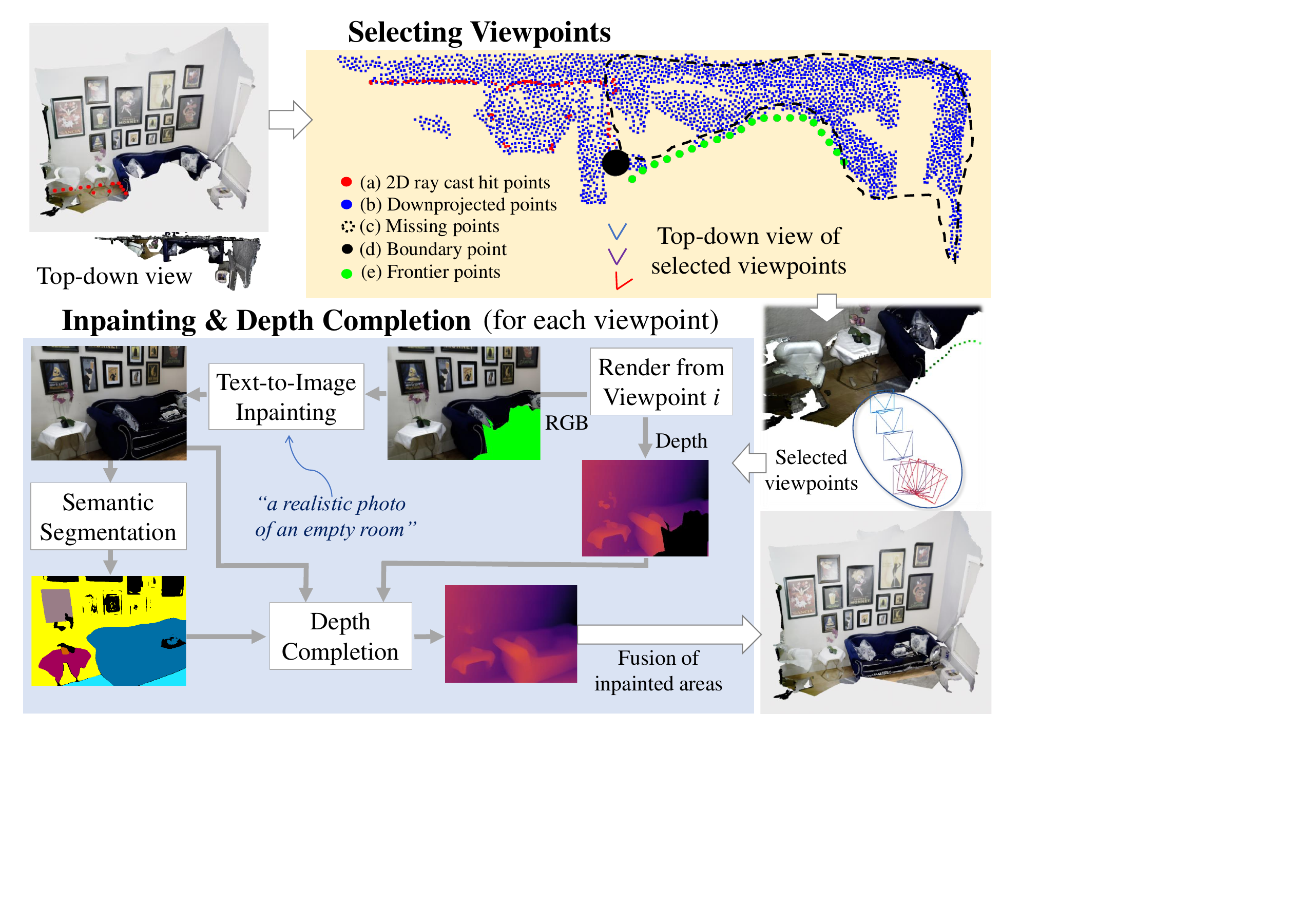}
    \caption{Given the 3D reconstruction, \ours{} first places virtual viewpoints whose union cover the frontier points. At each viewpoint, \ours{} renders an RGB-D image from the 3D points and performs image inpainting followed by depth completion with the help of the predicted semantics. Finally, it fuses the newly inpainted pixels back into 3D.}
    \label{fig:viewpoint_simple}
    \vspace{-15pt}
\end{figure}

To describe our strategy for placing virtual cameras, we define five types of point sets (see \figref{fig:viewpoint_simple}). a) The 2D ray cast hit points at the RVC height obtained from Section~\ref{sec:ray_cast}, b) \textit{downprojected points}: the reconstructed points projected onto a horizontal plane at the RVC height, c) \textit{missing points at the RVC height}: 
the subset of (b) whose vicinity does not include (a), d) \textit{boundary point}: the point in the largest cluster of (a) that is closest to (c), and e) \textit{frontier points}: which approximately represent the boundary of the area to be completed at the RVC height. 

We are interested in placing the frontier points in such a way that as the virtual cameras view these points, they gradually cover the missing part extending from the seen part. Therefore, we first find the boundary point that represents the boundary of the seen part and area to be completed. Next, to place the frontier points, we fit a concave hull~\cite{duckham2008efficient,moreira2007concave} on the downprojected points. Then, we sample points on the hull boundary starting from the point that is closest to the boundary point and extend towards the direction where there is no ray cast hit points for $2$m.

Finally, to generate the virtual camera trajectory, we search among all the frames in the video to find cameras that see the boundary point and pick the one with least pitch angle (i.e., camera looking down) as the first camera. Then, we move the camera back along its $z$-axis with a step size of $0.2$m until it sees half of the frontiers, and apply rotation along the floor normal direction with a step size of $10^{\circ}$ until the camera sees all the frontiers or it reaches the maximum rotation threshold ($30^{\circ}$). As shown in \figref{fig:viewpoint_simple}, these sequential viewpoints are later used in the point cloud completion module.

\textbf{Point cloud completion.} 
Given a target viewpoint, we first render an RGB-D image from the reconstructed point cloud using a differentiable point cloud renderer~\cite{chang2023pointersect}, along with a binary occupancy image. The rendered RGB image is taken as input to the Stable Diffusion inpainting model~\cite{rombach2021highresolution} to fill in the missing part indicated by the mask and a text prompt. In all our experiments we use the prompt \emph{``a realistic photo of an empty room"}, so as to discourage the inpainting model from hallucinating objects that do not exist in the actual scene. Next, we use a monocular depth estimation method, IronDepth~\cite{Bae2022}, on the inpainted RGB images for lifting the synthesized scene content to 3D. We backproject and fuse the inpainted areas into the reconstructed point cloud with the known camera parameters in an auto-regressive fashion. To better align the geometry of the new content to the reconstructed point cloud, we use the depth values rendered from the initial reconstruction as input to the depth estimation model and complete the rest of the pixels. Furthermore, we obtain a semantic segmentation of the inpainted image, and for the pixels predicted to be in the \emph{floor} class, we use their depths to the estimated floor plane as additional supervision to the depth completion module.

After inpainting new scene content from all the virtual viewpoints and completing the reconstructed point cloud, we again cast rays from the RVC height cameras to obtain the hit points as mentioned in Section~\ref{sec:ray_cast} and follow the same pose estimation method in Section~\ref{sec:pose_optim} to perform 2D point cloud registration with the LiDAR map.

\section{Experiments}

In this section we describe the set of experiments we conducted to evaluate \ours{}'s performance.

\textbf{Datasets.}
In our experiments, we use three real-world RGB-D datasets: Redwood~\cite{Park2017}, ScanNet~\cite{dai2017scannet} and a self-captured dataset from a large office. Both Redwood and ScanNet provide ground truth camera poses as well as the reconstructed meshes. We simulate the LiDAR maps by placing virtual sensors at the RVC height in the provided meshes and cast rays in all directions to collect hit points. 
To imitate the noise observed in real LiDAR scans, we further perturb the hit points with a 2D Gaussian with standard deviation of $1$cm and drop points with a probability of $10\%$. Note that the provided meshes from ScanNet contain missing parts and holes in the scene, therefore, the simulated LiDAR map is not perfect even before adding noise.
We use all apartment scenes from ScanNet and manually select several other scenes that are large and have better reconstructed meshes, which result in 36 scenes. For each scene, we randomly sample 6 videos from the provided camera stream. We include all scenes in the Redwood dataset and sample 20 videos from each scene. We additionally include 10 difficult videos for Redwood that capture mostly only the upper part of the scene. Among all the sampled videos, there are only 1 from ScanNet and 2 from Redwood that do not see the floor, which justify our assumption in Section~\ref{sec:ray_cast} that the floor is usually partially visible. 
We manually exclude these 3 videos. To collect data from the office scene, we mount a LiDAR on a ground robot and move it around the office to build the 2D map using GMapping~\cite{grisetti2007improved, grisetti2005improving}. We also record 5 video sequences at different locations in the office with a hand-held RGB-D camera. 

\textbf{Metrics.}
We report mean and median of the rotation and translation errors between the estimated pose registration and the ground truth. We denote $R_{\mu}, R_{med}$ as the mean and median of the estimated 2D rotation error in angles, and $T_{\mu}, T_{med}$ as the corresponding translation error in meters. Additionally, we report the success rate (SR) where a predicted registration is counted as successful if it has rotation error less than $10^\circ$ and translation error less than $0.3$m.

\subsection{Camera pose estimation on real-world RGB-D scans}\label{sec:exp_pose}
We present quantitative results for 2D pose estimation on both Redwood and ScanNet. 
Since there is no existing work that directly estimates the relative pose between a video and a 2D LiDAR point cloud, we perform a comparative study using varied versions of our proposed methods as baselines. The \textit{base method} denotes our method without the scene completion module.

\begin{figure}[b!]
    \centering
    \vspace{-5pt}
    \includegraphics[width=\linewidth]{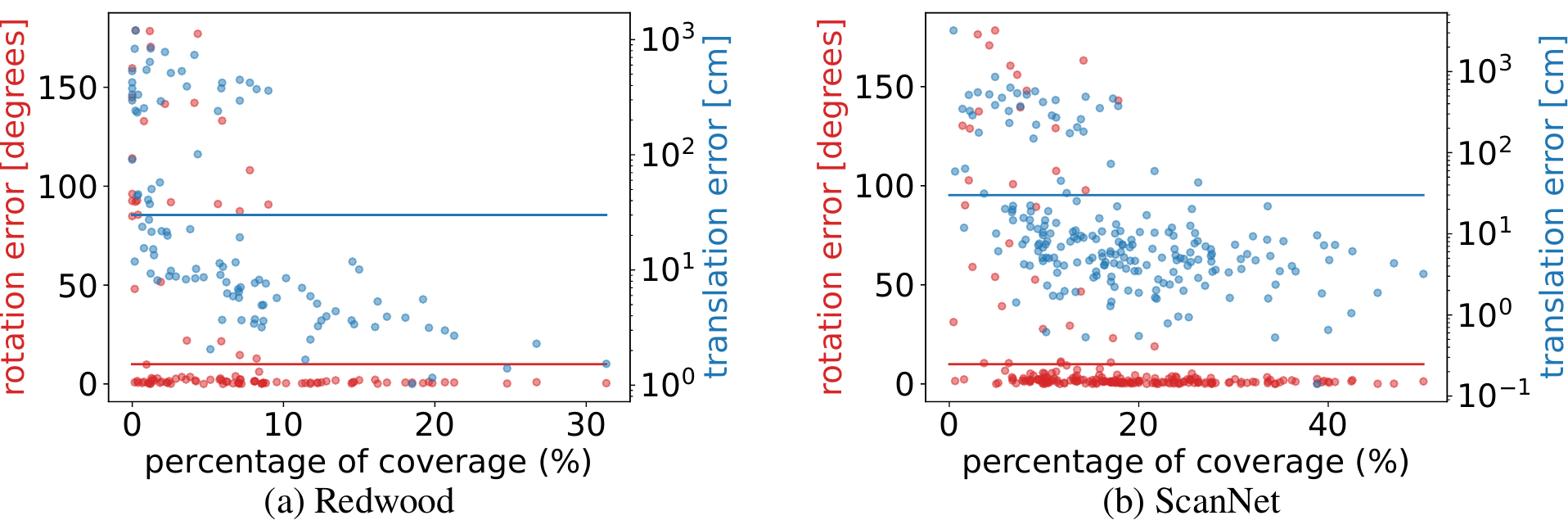}
    \caption{The effect of the reconstruction's coverage of the scene on the registration performance. The red line indicates the rotation error bound of $10^\circ$ and the blue line indicates the translation error bound of $0.3$m. It is shown that more data lies above the error bound as coverage decreases.}
    \label{fig:coverage_vs_registration}
    \vspace{-5pt}
\end{figure}

To verify our assumption that the missing points at the RVC height is one of the major failure causes, we investigate the correlation between the pose registration error and a \textit{coverage metric}. Specifically, for each sample, we align the ray cast points with the LiDAR map using the ground truth pose, then we measure the proportion of the LiDAR map covered by the ray cast points as the percentage of coverage. In \figref{fig:coverage_vs_registration}, we plot the pose estimation error of the base method as a function of this coverage metric. 
It is clear that as the coverage reduces, more data lie above the error bound. This motivates our scene completion module for filling in the missing data and improving the registration accuracy.

Next, we consider the full \ours{} pipeline on all the data by activating the scene completion module on every video. Indicated by ``\ours{}-all" in Table~\ref{table:complete}, we see that the scene completion module can hurt the pose registration performance if applied to all the videos. One major failure case is when the scene already has sufficient coverage at the RVC height, the completion module will be forced to generate geometry beyond the boundary of the observed scene. Therefore, the synthesized areas might not match the actual scene.


\begin{table}[t!]
\centering
\caption{The effect of the scene completion module based on different activating criteria. Base method is without the scene completion module. \ours{}-all applies the scene completion on all data and \ours{}-w/ gt. applies on the data that is considered failed measured with ground truth pose. \ours{} activates the completion module with the proposed decision criterion. }
\resizebox{\columnwidth}{!}{\begin{tabular}{cccccc}
\hline
\multicolumn{6}{c}{Redwood}                                                    \\
                                 & $R_{\mu}(^{\circ})\downarrow$    & $R_{med}(^{\circ})\downarrow$  & $T_{\mu}(m)\downarrow$  & $T_{med}(m)\downarrow$ & SR(\%)$\uparrow$  \\ \hline
\multicolumn{1}{c|}{Base method}  & 27.407 & 1.389  & 1.324 & 0.1    & 0.667   \\
\multicolumn{1}{c|}{\ours{}-all} &  27.934 & 1.695 & 1.264 & 0.148 & 0.648   \\
\multicolumn{1}{c|}{\ours{}}   & 18.787 & 0.975  & 0.794 & 0.09  & 0.741   \\ 
\multicolumn{1}{c|}{\ours{}-w/ gt.} & 15.829 & 1.12   & 0.646 & 0.084  & 0.806   \\
\hline
\multicolumn{6}{c}{ScanNet}                                                    \\
                                 & $R_{\mu}(^{\circ})\downarrow$    & $R_{med}(^{\circ})\downarrow$  & $T_{\mu}(m)\downarrow$  & $T_{med}(m)\downarrow$ & SR(\%)$\uparrow$  \\ \hline
\multicolumn{1}{c|}{Base method}  & 15.631 & 1.557  & 0.778 & 0.069  & 0.805   \\
\multicolumn{1}{c|}{\ours{}-all} &  29.203 & 2.862 & 0.86 & 0.111 & 0.660   \\
\multicolumn{1}{c|}{\ours{}}   &14.311  & 1.46   & 0.488 & 0.073   & 0.833  \\
\multicolumn{1}{c|}{\ours{}-w/ gt.} & 10.451 & 1.343  & 0.433 & 0.064  & 0.879 
\end{tabular}}
\label{table:complete}
\vspace{-15pt}
\end{table}

\begin{figure}[!t]
    \centering
    \includegraphics[width=\linewidth]{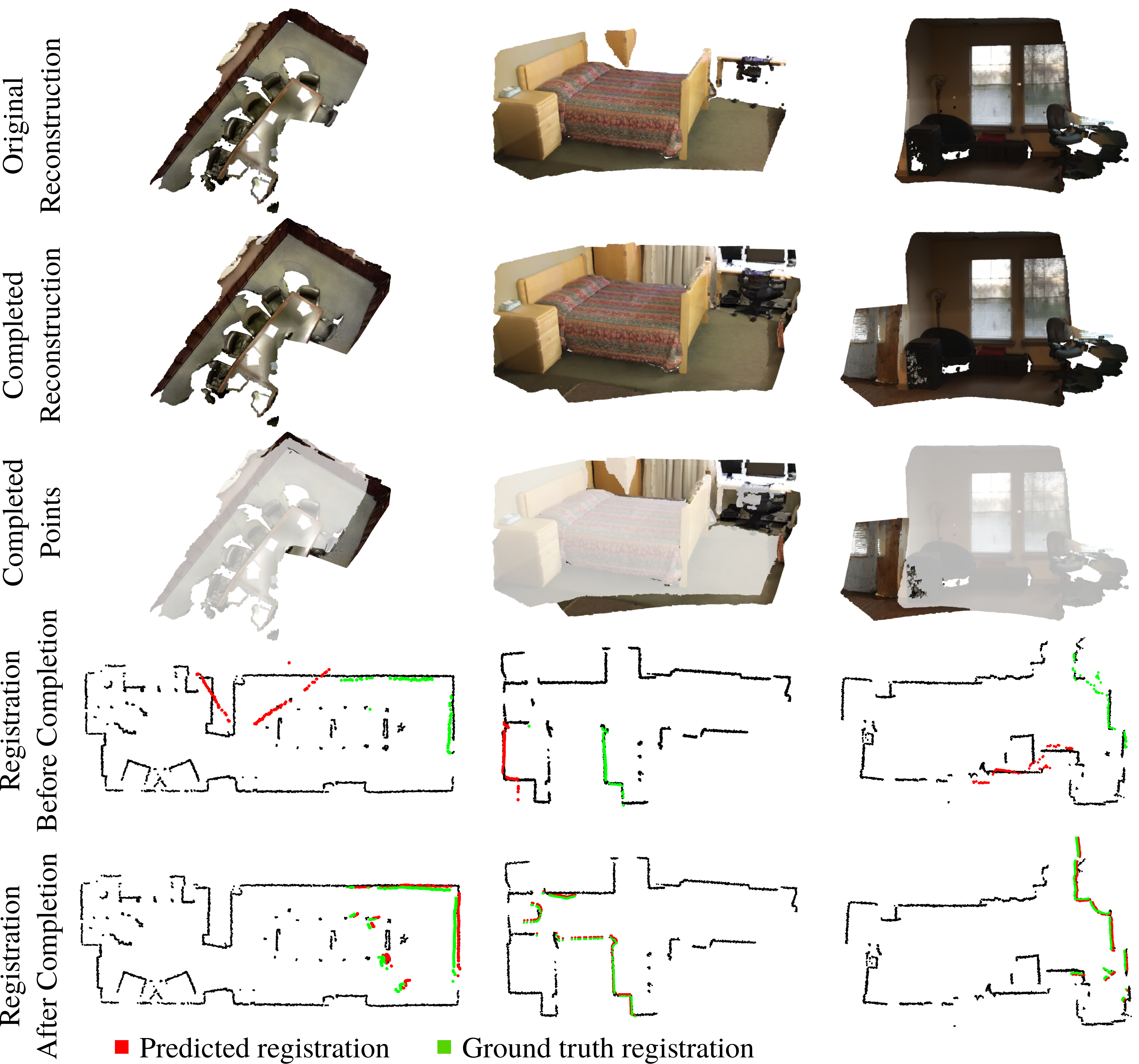}
    \caption{Qualitative results of \ours{} using and without using the point cloud completion module. We see that synthesizing the scene content at the RVC height improves the overall registration performance. The first two rows show the original and completed reconstructions, respectively, with the third row highlighting the added new points. In the last two rows, red points are the predicted and green ones are the ground truth registrations.}
    \label{fig:qualitative}
    \vspace{-15pt}
\end{figure}

To decide whether we should perform scene completion, we devised a decision criterion which effectively finds out if there are multiple local minima among the optimized poses in Section~\ref{sec:pose_optim}. Specifically, we consider all poses and check if there are two poses which have small loss values and at the same time, are not close. 
We define $\mymat{T}_i$ and $\mymat{T}_j$ to be close if the relative rotation is smaller than $\theta_R$ and relative translation is smaller than $\theta_T$. More concretely, we start with the pose $\mymat{T}_1$ with the smallest loss value after optimization. We then remove its neighbors which are close from consideration and pick the second best pose $\mymat{T}_2$. Let $\mymat{L}_1$ and $\mymat{L}_2$ be the loss values associated with $\mymat{T}_1$ and $\mymat{T}_2$ respectively. If these two poses have similar loss values (i.e. $|\mymat{L}_1-\mymat{L}_2|<c$), then it means there are multiple local minima, therefore, scene completion is needed.
To determine the values of $\theta_R, \theta_T, c$, we use $20\%$ of data from ScanNet and perform parameter search. We select the parameter set that allows \ours{} to have the minimum rotation error on this $20\%$ of the data. Finally, we use $\theta_R=20^\circ$, $\theta_T=0.3$m and $c=20$ for \ours{} on all data.
As shown in Table~\ref{table:complete}, this activation criterion captures the cases that require completion well and therefore \ours{} performs better than the base method on both datasets. \figref{fig:qualitative} shows qualitatively how scene completion in \ours{} helps pose registration on the scenes where the base method fails.

To verify the performance of the designed decision criterion for the completion module, we further conduct an experiment that activates the point cloud completion module using the pose registration error measured with ground truth, which represents the upper bound of \ours{}'s performance. ``\ours{}-w/ gt." in Table~\ref{table:view_compare} shows that our designed decision criterion allows \ours{} to perform close to its upper bound, and with a better developed criterion, the performance of \ours{} can be further improved.

\textbf{Floor map augmentation with \ours{}.}
In addition to public datasets for which we simulated the LiDAR scans, we present qualitative results on the self-captured office scan with a noisy, real-world LiDAR map. We demonstrate that \ours{} can help augment the LiDAR map after pose registration. As shown in \figref{fig:office-scan}, 4 videos recorded separately are registered to the same LiDAR scan to achieve 3D reconstruction of a large scene. Furthermore, we show that semantics information can be added to the LiDAR map, which is useful for enabling more informative floor layouts.


\subsection{Comparison of strategies for target viewpoint selection} \label{sec:compare_viewpoint}
We observe that the inpainting model is sensitive to how we select the novel viewpoints for rendering RGB images and for performing inpainting.
Therefore, we investigate the effect of different viewpoint selection strategies. In Section~\ref{sec:ray_cast}, we move the cameras down to the RVC height to cast rays for emulating LiDAR hit points. One idea is to directly use these virtual cameras at the RVC height to maximize the amount of new content. Another approach is to move the SLAM camera poses back by $0.5$m along the camera's $z$-axis to increase the field of view while guaranteeing to keep a certain portion of the seen content. We compare these two additional view selection strategies with \ours{} as mentioned in Section~\ref{sec:view_select_complete}. As shown in Table~\ref{table:view_compare}, \ours{} outperforms these two baseline viewpoint selection strategies. We observe that directly inpainting from the viewpoints at the RVC height gives the most information for the 2D hit points. However, sometimes these viewing angles are much different from the original camera poses, and therefore are relying on the inpainting model to synthesize a large empty space without enough information for grounding to the actual scene. Stepping back from the original camera poses is a way to guarantee that the input RGB images to the inpainting model retains certain amount of content. However, naively moving the cameras back can result in undesired viewpoints or viewpoints that do not help with filling in missing area at the RVC height. 

In contrast, \ours{}'s viewpoint selection strategy allows the inpainting model to start from already seen parts and move toward the missing part at the RVC height with gradual viewpoint change. As shown in \figref{fig:qualitative}, \ours{} is able to fill in reasonable geometry at the RVC height. Note that \ours{}'s main goal is to complete the points for the purpose of pose registration, instead of generating the whole scene. Therefore, although the completed area does not cover a large space, it can improve the registration performance as long as the missing areas at the RVC height are reconstructed well.

\begin{table}[t!]
\centering
\caption{The effect of different viewpoint selection strategies for scene completion on pose registration performance. \textit{Step back} moves SLAM camera poses back by $0.5$m along the camera's $z$-axis. \textit{RVC} projects the viewpoints down to the RVC height from the SLAM poses.}
\resizebox{\columnwidth}{!}{\begin{tabular}{cccccc}
\hline
\multicolumn{6}{c}{Redwood}                                                 \\
                               & $R_{\mu}(^{\circ})\downarrow$    & $R_{med}(^{\circ})\downarrow$  & $T_{\mu}(m)\downarrow$  & $T_{med}(m)\downarrow$ & SR(\%)$\uparrow$  \\ \hline
\multicolumn{1}{c|}{Step back} & 27.179 & 1.467  & 1.303 & 0.102  & 0.685   \\
\multicolumn{1}{c|}{RVC}       & 29.32  & 1.306  & 1.097 & 0.096  & 0.694   \\
\multicolumn{1}{c|}{\ours{}}  & \textbf{18.787} & \textbf{0.975}  & \textbf{0.794} & \textbf{0.09}  & \textbf{0.741}   \\ \hline
\multicolumn{6}{c}{ScanNet}                                                 \\
                               & $R_{\mu}(^{\circ})\downarrow$    & $R_{med}(^{\circ})\downarrow$  & $T_{\mu}(m)\downarrow$  & $T_{med}(m)\downarrow$ & SR(\%)$\uparrow$  \\ \hline
\multicolumn{1}{c|}{Step back} & 16.076 & 1.63   & 0.645 & 0.073  & 0.833   \\
\multicolumn{1}{c|}{RVC}       & 15.082 & 1.458  & 0.657 & 0.073  & 0.814   \\
\multicolumn{1}{c|}{\ours{}}  & \textbf{14.311}  & \textbf{1.46}   & \textbf{0.488} & \textbf{0.073}   & \textbf{0.833} 
\end{tabular}}
\label{table:view_compare}
\vspace{-15pt}
\end{table}



\section{Conclusion}

We presented \ours{} for aligning RGB-D videos to 2D LiDAR maps obtained by RVCs. After building a 3D semantic map using the image sequence, \ours{} performs 2D ray casting to emulate the LiDAR measurements, and inputs the hit points to our pose optimization module for registering to the LiDAR map. 
Our key observation is that the registration error correlates with the missing points at the RVC height. To fill in the missing information, we introduced a scene completion technique that leverages a pre-trained text-to-image model paired with a novel viewpoint selection strategy.
We show that after using the completed point clouds, our registration module can solve up to 20\% of the instances where it has failed before. In addition, we demonstrate that our method can be used to align short sequences of videos collected independently from different rooms, which enables augmenting LiDAR maps with 3D geometry and semantics.

While \ours{} shows promising results for video to LiDAR alignment, it has some limitations that need to be addressed. First, we assume that a portion of the floor is visible in the video, since our method relies on estimating the floor plane. One way to eliminate this assumption is to use additional sensing such as IMUs to estimate the gravity direction, then performing ray casting at multiple heights as possible inputs for registration.
Moreover, we note that using pre-trained inpainting and depth completion models can generate implausible geometries that may be detrimental to the pose registration performance. Although our strategy for selecting viewpoints mitigates this issue, we see few instances impacted by inaccurate completions of the scene, which leaves an exciting open problem for future studies.

\clearpage
\bibliographystyle{unsrt}
\bibliography{ref}

\begin{thebibliography}{10}

\bibitem{yu2020monocular}
Huai Yu, Weikun Zhen, Wen Yang, Ji~Zhang, and Sebastian Scherer.
\newblock Monocular camera localization in prior lidar maps with {2D-3D} line correspondences.
\newblock In {\em 2020 IEEE/RSJ International Conference on Intelligent Robots and Systems (IROS)}, pages 4588--4594. IEEE, 2020.

\bibitem{Feng2019ICRA}
Mengdan Feng, Sixing Hu, Marcelo Ang, and Gim~Hee Lee.
\newblock {2D3D-MatchNet}s: Learning to match keypoints across 2d image and 3d point cloud.
\newblock In {\em The IEEE International Conference on Robotics and Automation (ICRA)}, May 2019.

\bibitem{li2021deepi2p}
Jiaxin Li and Gim~Hee Lee.
\newblock {DeepI2P}: Image-to-point cloud registration via deep classification.
\newblock In {\em Proceedings of the IEEE/CVF Conference on Computer Vision and Pattern Recognition}, pages 15960--15969, 2021.

\bibitem{ren2022corri2p}
Siyu Ren, Yiming Zeng, Junhui Hou, and Xiaodong Chen.
\newblock {CorrI2P}: Deep image-to-point cloud registration via dense correspondence.
\newblock {\em IEEE Transactions on Circuits and Systems for Video Technology}, 33(3):1198--1208, 2022.

\bibitem{miao2023poses}
Jinyu Miao, Kun Jiang, Yunlong Wang, Tuopu Wen, Zhongyang Xiao, Zheng Fu, Mengmeng Yang, Maolin Liu, and Diange Yang.
\newblock Poses as queries: Image-to-lidar map localization with transformers.
\newblock {\em arXiv preprint arXiv:2305.04298}, 2023.

\bibitem{jeon2022efghnet}
Yurim Jeon and Seung-Woo Seo.
\newblock {EFGHNet}: A versatile image-to-point cloud registration network for extreme outdoor environment.
\newblock {\em IEEE Robotics and Automation Letters}, 7(3):7511--7517, 2022.

\bibitem{CHEN2022209}
Kuangyi Chen, Huai Yu, Wen Yang, Lei Yu, Sebastian Scherer, and Gui-Song Xia.
\newblock {I2D-Loc: Camera Localization via Image to LiDAR Depth Flow}.
\newblock In {\em ISPRS Journal of Photogrammetry and Remote Sensing}, volume 194, pages 209--221, 2022.

\bibitem{wijmans2017exploiting}
Erik Wijmans and Yasutaka Furukawa.
\newblock Exploiting 2d floorplan for building-scale panorama rgbd alignment.
\newblock In {\em Proceedings of the IEEE Conference on Computer Vision and Pattern Recognition}, pages 308--316, 2017.

\bibitem{sokolova2022iros}
Anna Sokolova, Filipp Nikitin, Anna Vorontsova, and Anton Konushin.
\newblock Floorplan-aware camera poses refinement.
\newblock In {\em 2022 IEEE/RSJ International Conference on Intelligent Robots and Systems (IROS)}, pages 4857--4864, 2022.

\bibitem{besl1992method}
Paul~J Besl and Neil~D McKay.
\newblock Method for registration of {3-D} shapes.
\newblock In {\em Sensor fusion IV: control paradigms and data structures}, volume 1611, pages 586--606. Spie, 1992.

\bibitem{segal2009generalized}
Aleksandr Segal, Dirk Haehnel, and Sebastian Thrun.
\newblock {Generalized-ICP}.
\newblock In {\em Robotics: science and systems}, volume~2, page 435. Seattle, WA, 2009.

\bibitem{zhou2016fast}
Qian-Yi Zhou, Jaesik Park, and Vladlen Koltun.
\newblock Fast global registration.
\newblock In {\em European conference on computer vision}, pages 766--782. Springer, 2016.

\bibitem{Yang20tro-teaser}
H.~Yang, J.~Shi, and L.~Carlone.
\newblock {TEASER: Fast and Certifiable Point Cloud Registration}.
\newblock {\em {IEEE} Trans. Robotics}, 2020.

\bibitem{yang2019polynomial}
Heng Yang and Luca Carlone.
\newblock A polynomial-time solution for robust registration with extreme outlier rates.
\newblock {\em arXiv preprint arXiv:1903.08588}, 2019.

\bibitem{rusu2008aligning}
Radu~Bogdan Rusu, Nico Blodow, Zoltan~Csaba Marton, and Michael Beetz.
\newblock Aligning point cloud views using persistent feature histograms.
\newblock In {\em 2008 IEEE/RSJ international conference on intelligent robots and systems}, pages 3384--3391. IEEE, 2008.

\bibitem{rusu2009fast}
Radu~Bogdan Rusu, Nico Blodow, and Michael Beetz.
\newblock Fast point feature histograms ({FPFH}) for {3D} registration.
\newblock In {\em 2009 IEEE international conference on robotics and automation}, pages 3212--3217. IEEE, 2009.

\bibitem{chao2023category}
Jun-Jee Chao, Selim Engin, Nicolai H{\"a}ni, and Volkan Isler.
\newblock Category-level global camera pose estimation with multi-hypothesis point cloud correspondences.
\newblock In {\em 2023 IEEE International Conference on Robotics and Automation (ICRA)}, pages 3800--3807. IEEE, 2023.

\bibitem{deng2018ppfnet}
Haowen Deng, Tolga Birdal, and Slobodan Ilic.
\newblock {PPFNet}: Global context aware local features for robust {3D} point matching.
\newblock In {\em Proceedings of the IEEE conference on computer vision and pattern recognition}, pages 195--205, 2018.

\bibitem{zeng20173dmatch}
Andy Zeng, Shuran Song, Matthias Nie{\ss}ner, Matthew Fisher, Jianxiong Xiao, and Thomas Funkhouser.
\newblock {3DMatch}: Learning local geometric descriptors from {RGB-D} reconstructions.
\newblock In {\em Proceedings of the IEEE conference on computer vision and pattern recognition}, pages 1802--1811, 2017.

\bibitem{gojcic2019perfect}
Zan Gojcic, Caifa Zhou, Jan~D Wegner, and Andreas Wieser.
\newblock The perfect match: {S3D} point cloud matching with smoothed densities.
\newblock In {\em Proceedings of the IEEE/CVF conference on computer vision and pattern recognition}, pages 5545--5554, 2019.

\bibitem{wang2019deep}
Yue Wang and Justin~M Solomon.
\newblock Deep closest point: Learning representations for point cloud registration.
\newblock In {\em Proceedings of the IEEE/CVF international conference on computer vision}, pages 3523--3532, 2019.

\bibitem{ginzburg2022deep}
Dvir Ginzburg and Dan Raviv.
\newblock Deep confidence guided distance for 3d partial shape registration.
\newblock {\em arXiv preprint arXiv:2201.11379}, 2022.

\bibitem{qi2017pointnet++}
Charles~Ruizhongtai Qi, Li~Yi, Hao Su, and Leonidas~J Guibas.
\newblock {PointNet++}: Deep hierarchical feature learning on point sets in a metric space.
\newblock {\em Advances in neural information processing systems}, 30, 2017.

\bibitem{yaoki2019pointnetlk}
Yasuhiro Aoki, Hunter Goforth, Rangaprasad Arun~Srivatsan, and Simon Lucey.
\newblock {PointNetLK}: Robust \& efficient point cloud registration using {PointNet}.
\newblock In {\em The IEEE Conference on Computer Vision and Pattern Recognition (CVPR)}, June 2019.

\bibitem{wang2019prnet}
Yue Wang and Justin~M Solomon.
\newblock {PRNet}: Self-supervised learning for partial-to-partial registration.
\newblock {\em Advances in neural information processing systems}, 32, 2019.

\bibitem{yew2020-RPMNet}
Zi~Jian Yew and Gim~Hee Lee.
\newblock {RPM-Net}: Robust point matching using learned features.
\newblock In {\em Conference on Computer Vision and Pattern Recognition (CVPR)}, 2020.

\bibitem{Fu2021RGM}
Kexue Fu, Shaolei Liu, Xiaoyuan Luo, and Manning Wang.
\newblock Robust point cloud registration framework based on deep graph matching.
\newblock {\em Internaltional Conference on Computer Vision and Pattern Recogintion (CVPR)}, 2021.

\bibitem{jiang2021sampling}
Haobo Jiang, Yaqi Shen, Jin Xie, Jun Li, Jianjun Qian, and Jian Yang.
\newblock Sampling network guided cross-entropy method for unsupervised point cloud registration.
\newblock In {\em Proceedings of the IEEE/CVF International Conference on Computer Vision}, pages 6128--6137, 2021.

\bibitem{wu2021feature}
Bingli Wu, Jie Ma, Gaojie Chen, and Pei An.
\newblock Feature interactive representation for point cloud registration.
\newblock In {\em Proceedings of the IEEE/CVF International Conference on Computer Vision}, pages 5530--5539, 2021.

\bibitem{idam}
Jiahao Li, Changhao Zhang, Ziyao Xu, Hangning Zhou, and Chi Zhang.
\newblock Iterative distance-aware similarity matrix convolution with mutual-supervised point elimination for efficient point cloud registration.
\newblock In {\em European Conference on Computer Vision (ECCV)}, 2020.

\bibitem{lee2021deeppro}
Donghoon Lee, Onur~C Hamsici, Steven Feng, Prachee Sharma, and Thorsten Gernoth.
\newblock {DeepPRO}: Deep partial point cloud registration of objects.
\newblock In {\em Proceedings of the IEEE/CVF International Conference on Computer Vision}, pages 5683--5692, 2021.

\bibitem{predator}
Shengyu Huang, Zan Gojcic, Mikhail Usvyatsov, and Konrad~Schindler Andreas~Wieser.
\newblock {PREDATOR}: Registration of 3d point clouds with low overlap.
\newblock In {\em IEEE Conference on Computer Vision and Pattern Recognition, CVPR}, 2021.

\bibitem{choy2020deep}
Christopher Choy, Wei Dong, and Vladlen Koltun.
\newblock Deep global registration.
\newblock In {\em Proceedings of the IEEE/CVF conference on computer vision and pattern recognition}, pages 2514--2523, 2020.

\bibitem{qin2022geometric}
Zheng Qin, Hao Yu, Changjian Wang, Yulan Guo, Yuxing Peng, and Kai Xu.
\newblock Geometric transformer for fast and robust point cloud registration.
\newblock In {\em Proceedings of the IEEE/CVF Conference on Computer Vision and Pattern Recognition (CVPR)}, pages 11143--11152, June 2022.

\bibitem{zhang2021view}
Xuancheng Zhang, Yutong Feng, Siqi Li, Changqing Zou, Hai Wan, Xibin Zhao, Yandong Guo, and Yue Gao.
\newblock View-guided point cloud completion.
\newblock In {\em Proceedings of the IEEE/CVF Conference on Computer Vision and Pattern Recognition}, pages 15890--15899, 2021.

\bibitem{yu2021pointr}
Xumin Yu, Yongming Rao, Ziyi Wang, Zuyan Liu, Jiwen Lu, and Jie Zhou.
\newblock {PoinTr}: Diverse point cloud completion with geometry-aware transformers.
\newblock In {\em ICCV}, 2021.

\bibitem{li2021spgan}
Ruihui Li, Xianzhi Li, Ke-Hei Hui, and Chi-Wing Fu.
\newblock {SP-GAN}:sphere-guided 3d shape generation and manipulation.
\newblock {\em ACM Transactions on Graphics (Proc. SIGGRAPH)}, 40(4), 2021.

\bibitem{yan2022shapeformer}
Xingguang Yan, Liqiang Lin, Niloy~J. Mitra, Dani Lischinski, Danny Cohen-Or, and Hui Huang.
\newblock {ShapeFormer}: Transformer-based shape completion via sparse representation.
\newblock In {\em Proceedings of the IEEE/CVF Conference on Computer Vision and Pattern Recognition}, 2022.

\bibitem{hani20233d}
Nicolai H{\"a}ni, Jun-Jee Chao, and Volkan Isler.
\newblock {3D} surface reconstruction in the wild by deforming shape priors from synthetic data.
\newblock {\em arXiv preprint arXiv:2302.12883}, 2023.

\bibitem{autosdf2022}
Paritosh Mittal, Yen-Chi Cheng, Maneesh Singh, and Shubham Tulsiani.
\newblock {AutoSDF}: Shape priors for 3d completion, reconstruction and generation.
\newblock In {\em CVPR}, 2022.

\bibitem{li2023voxformer}
Yiming Li, Zhiding Yu, Christopher Choy, Chaowei Xiao, Jose~M Alvarez, Sanja Fidler, Chen Feng, and Anima Anandkumar.
\newblock {VoxFormer}: Sparse voxel transformer for camera-based 3d semantic scene completion.
\newblock In {\em Proceedings of the IEEE/CVF Conference on Computer Vision and Pattern Recognition (CVPR)}, 2023.

\bibitem{xia2023scpnet}
Zhaoyang Xia, Youquan Liu, Xin Li, Xinge Zhu, Yuexin Ma, Yikang Li, Yuenan Hou, and Yu~Qiao.
\newblock {SCPNet}: Semantic scene completion on point cloud.
\newblock In {\em Proceedings of the IEEE/CVF Conference on Computer Vision and Pattern Recognition}, pages 17642--17651, 2023.

\bibitem{song2017semantic}
Shuran Song, Fisher Yu, Andy Zeng, Angel~X Chang, Manolis Savva, and Thomas Funkhouser.
\newblock Semantic scene completion from a single depth image.
\newblock In {\em Proceedings of the IEEE conference on computer vision and pattern recognition}, pages 1746--1754, 2017.

\bibitem{cai2021semantic}
Yingjie Cai, Xuesong Chen, Chao Zhang, Kwan-Yee Lin, Xiaogang Wang, and Hongsheng Li.
\newblock Semantic scene completion via integrating instances and scene in-the-loop.
\newblock In {\em Proceedings of the IEEE/CVF Conference on Computer Vision and Pattern Recognition}, pages 324--333, 2021.

\bibitem{cao2022monoscene}
Anh-Quan Cao and Raoul de~Charette.
\newblock Monoscene: Monocular 3d semantic scene completion.
\newblock In {\em Proceedings of the IEEE/CVF Conference on Computer Vision and Pattern Recognition}, pages 3991--4001, 2022.

\bibitem{ha2022semabs}
Huy Ha and Shuran Song.
\newblock Semantic abstraction: Open-world 3{D} scene understanding from 2{D} vision-language models.
\newblock In {\em Proceedings of the 2022 Conference on Robot Learning}, 2022.

\bibitem{wang2023semantic}
Fengyun Wang, Dong Zhang, Hanwang Zhang, Jinhui Tang, and Qianru Sun.
\newblock Semantic scene completion with cleaner self.
\newblock In {\em Proceedings of the IEEE/CVF Conference on Computer Vision and Pattern Recognition}, pages 867--877, 2023.

\bibitem{hoellein2023text2room}
Lukas H{\"o}llein, Ang Cao, Andrew Owens, Justin Johnson, and Matthias Nie{\ss}ner.
\newblock Text2room: Extracting textured 3d meshes from 2d text-to-image models, 2023.

\bibitem{SceneScape}
Rafail Fridman, Amit Abecasis, Yoni Kasten, and Tali Dekel.
\newblock {SceneScape}: Text-driven consistent scene generation.
\newblock {\em arXiv preprint arXiv:2302.01133}, 2023.

\bibitem{teed2021droid}
Zachary Teed and Jia Deng.
\newblock {DROID-SLAM: Deep Visual SLAM for Monocular, Stereo, and RGB-D Cameras}.
\newblock {\em Advances in neural information processing systems}, 2021.

\bibitem{choi2015robust}
Sungjoon Choi, Qian-Yi Zhou, and Vladlen Koltun.
\newblock Robust reconstruction of indoor scenes.
\newblock In {\em Proceedings of the IEEE conference on computer vision and pattern recognition}, pages 5556--5565, 2015.

\bibitem{park2017colored}
Jaesik Park, Qian-Yi Zhou, and Vladlen Koltun.
\newblock Colored point cloud registration revisited.
\newblock In {\em Proceedings of the IEEE international conference on computer vision}, pages 143--152, 2017.

\bibitem{siddiqui2023panoptic}
Yawar Siddiqui, Lorenzo Porzi, Samuel~Rota Bul{\`o}, Norman M{\"u}ller, Matthias Nie{\ss}ner, Angela Dai, and Peter Kontschieder.
\newblock Panoptic lifting for {D3D} scene understanding with neural fields.
\newblock In {\em Proceedings of the IEEE/CVF Conference on Computer Vision and Pattern Recognition}, pages 9043--9052, 2023.

\bibitem{cheng2021mask2former}
Bowen Cheng, Ishan Misra, Alexander~G. Schwing, Alexander Kirillov, and Rohit Girdhar.
\newblock Masked-attention mask transformer for universal image segmentation.
\newblock In {\em Proceedings of the IEEE/CVF Conference on Computer Vision and Pattern Recognition}, 2022.

\bibitem{duckham2008efficient}
Matt Duckham, Lars Kulik, Mike Worboys, and Antony Galton.
\newblock Efficient generation of simple polygons for characterizing the shape of a set of points in the plane.
\newblock {\em Pattern recognition}, 41(10):3224--3236, 2008.

\bibitem{moreira2007concave}
Adriano Moreira and Maribel~Yasmina Santos.
\newblock Concave hull: A k-nearest neighbours approach for the computation of the region occupied by a set of points.
\newblock 2007.

\bibitem{chang2023pointersect}
Jen-Hao~Rick Chang, Wei-Yu Chen, Anurag Ranjan, Kwang~Moo Yi, and Oncel Tuzel.
\newblock Pointersect: Neural rendering with cloud-ray intersection.
\newblock In {\em Proceedings of the IEEE Conference on Computer Vision and Pattern Recognition}, 2023.

\bibitem{rombach2021highresolution}
Robin Rombach, Andreas Blattmann, Dominik Lorenz, Patrick Esser, and Bj{\"o}rn Ommer.
\newblock High-resolution image synthesis with latent diffusion models.
\newblock In {\em Proceedings of the IEEE/CVF conference on computer vision and pattern recognition}, pages 10684--10695, 2022.

\bibitem{Bae2022}
Gwangbin Bae, Ignas Budvytis, and Roberto Cipolla.
\newblock {IronDepth}: Iterative refinement of single-view depth using surface normal and its uncertainty.
\newblock In {\em British Machine Vision Conference (BMVC)}, 2022.

\bibitem{Park2017}
Jaesik Park, Qian-Yi Zhou, and Vladlen Koltun.
\newblock Colored point cloud registration revisited.
\newblock In {\em ICCV}, 2017.

\bibitem{dai2017scannet}
Angela Dai, Angel~X. Chang, Manolis Savva, Maciej Halber, Thomas Funkhouser, and Matthias Nie{\ss}ner.
\newblock {ScanNet}: Richly-annotated 3d reconstructions of indoor scenes.
\newblock In {\em Proc. Computer Vision and Pattern Recognition (CVPR), IEEE}, 2017.

\bibitem{grisetti2007improved}
Giorgio Grisetti, Cyrill Stachniss, and Wolfram Burgard.
\newblock Improved techniques for grid mapping with rao-blackwellized particle filters.
\newblock {\em IEEE transactions on Robotics}, 23(1):34--46, 2007.

\bibitem{grisetti2005improving}
Giorgio Grisetti, Cyrill Stachniss, and Wolfram Burgard.
\newblock Improving grid-based slam with rao-blackwellized particle filters by adaptive proposals and selective resampling.
\newblock In {\em Proceedings of the 2005 IEEE international conference on robotics and automation}, pages 2432--2437. IEEE, 2005.

\end{thebibliography}

\end{document}